\documentclass[
]{ceurart}

\usepackage[center]{caption}
\usepackage{amsmath}
\usepackage{textalpha}

\begin{document}

\copyrightyear{2020}
\copyrightclause{Copyright for this paper by its authors.
  Use permitted under Creative Commons License Attribution 4.0
  International (CC BY 4.0).}

\conference{FIRE'20:}

\title{NUIG-Shubhanker@Dravidian-CodeMix-FIRE2020: Sentiment Analysis of Code-Mixed Dravidian text using XLNet}

\author[1]{Shubhanker Banerjee}[%
email=S.Banerjee3@nuigalway.ie,
]
\address[1]{National University Of Ireland Galway, Ireland}

\author[2]{Arun Jayapal}[%
email=jayapala@tcd.ie,
]
\address[2]{Trinity college Dublin, Ireland}

\author[3]{Sajeetha Thavareesan}[%
orcid=0000-0002-6252-5393,
email=sajeethas@esn.ac.lk
]
\address[3]{Eastern University, Sri Lanka}

\begin{abstract}
Social media has penetrated into multi-lingual societies, however most of them use English to be a preferred language for communication. So it looks natural for them to mix their cultural language with English during conversations resulting in abundance of multilingual data -- call this {\em code-mixed data}, available in today's world. Downstream NLP tasks using such data is challenging due to the semantic nature of it being spread across multiple languages. One such NLP task is Sentiment analysis; for this we use an auto-regressive XLNet model to perform sentiment analysis on code-mixed Tamil-English and Malayalam-English datasets.
\end{abstract}


\maketitle

\section{Introduction}
Social media content results in large data feeds from wide geographies. Since multiple geographies are involved, the data is multilingual in nature resulting in {\em code mixing} \footnote{Code mixing refers to the linguistic units from different languages being used together by multilingual users} often. Sentiment analysis on code-mixed text allows to gain insights on the trends prevalent in different geographies however is a challenge due to the non-trivial nature involved in inferring the semantics of such data. In this paper, we address these challenges using XLNet\cite{yang2019xlnet}\footnote{XLNet is an auto-regressive model\cite{gregor2014deep} which is built with the transformer architecture\cite{vaswani2017attention} with a two stream attention mechanism\cite{bahdanau2014neural}. The two stream attention mechanism ensures that the language model obtained through training can predict missing words on the basis of bidirectional context. Bidirectional context in XLNet is achieved by permutation language modelling.} framework. We have fine tuned the pre-trained XLNet model with the available data without any additional pre-processing mechanisms. The rest of the paper is organized as follows. Section 2 illustrates the related work done in the field. Section 3 explains the dataset and the task. Section 4 demonstrates the approach we used and briefly explains the architecture of XLNet. Results are discussed in Section 5. 
\section{Related Work}
Multilingual users have the tendency to mix linguistic units in the social media resulting in code-mixed data being easily available. The phenomenon of code-mixing is explained in \cite{kim2006reasons, chakravarthi-etal-2019-wordnet,chakravarthi-etal-2019-multilingual,chakravarthi2019comparison,chakravarthi2020survey,chakravarthi2020leveraging} and provides an analysis on the possible reasons behind code-mixing. This is done by identifying the languages involved in the code-mixed data which looks inevitable. In the past, several approaches were taken and experiments were conducted aimed at the detection of languages in code-mixed data \cite{unknown}\cite{inproceedings}\cite{10.1017/S1351324910000252}. A review of many research works on code-mixing is discussed in \cite{jose2020survey}.

\subsection{Code-mixed data} Since code-mixed is mostly sourced from social media platforms, the data in it's raw form is highly unstructured and hence corpus creation to organize this unstructured data into datasets for further analysis pose a challenge. For some of the Indian languages, \cite{chakravarthi-etal-2020-corpus} has compiled a Tamil-English code-mixed dataset, the first annotated Tanglish\footnote{Tanglish refers to code switching between Tamil and English, a term predominantly used in the Tamil community} dataset. Similarly, \cite{chakravarthi-etal-2020-sentiment} published a dataset for Malayalam-English code-mixed data, where the authors also provided references to the availability of other code-mixed datasets such as Chinese-English and Spanish-English. But significant work hasn't been done in the area of corpus creation for code-mixing for Indian languages. The Indian languages are considered to be under-resourced and so there is less interest in performing NLP tasks on these languages. 

\subsection{Sentiment-analysis}
Sentiment analysis is a well known NLP task that infers the positive, negative and neutral sentiments from a statement in question. However there are very few works on the sentiment analysis over code-mixed data; \cite{dravidiansentiment-ceur}\cite{dravidiansentiment-acm} provides an overview of the work done on sentiment analysis of Dravidian code-mixed text. Another work, \cite{icac3n-20} compares the performance of different transformer architectures on the task of sentiment analysis of code-mixed data. 
\cite{7275819} employed an approach based on lexicon to assign sentiment to Hindi-English code-mixed text. \cite{rani-etal-2020-comparative} illustrates a method to detect hate speech in code-mixed Hinglish dataset. For the purpose of conducting this research they used FIRE 2013 and FIRE 2014 datasets. \cite{article} used a LSTM \cite{hochreiter1997long} based approach to improve the state-of-art performance \cite{7275819} on the hinglish datasets by 18 percent. \cite{choudhary2018sentiment} used shared parameters in a siamese network \cite{roy2019siamese} to project the code-mixed sentences and sentences in standard languages into a common sentiment space. The similarity of projected sentences is an indicative of how similar their sentiments are, similar sentences have similar sentiment. Ensemble based techniques have also been used for sentiment analysis of code-mixed data, \cite{jhanwar2018ensemble} proposed an ensemble model of a character-trigrams based LSTM and a word-ngrams naive bayes to detect the sentiment in Hindi-english code-mixed data. \cite{ghosh2017sentiment} have used a multilayer perceptron to perform sentiment analysis on code-mixed data extracted from social media platforms. \cite{kumar2020baksa} used an ensemble of a convolutional neural network and a self-attention based LSTM for sentiment analysis of Spanglish and Hinglish text.

\section{Dataset for sentiment analysis}

In spoken and written conversations, it is observed that the usage of lexicon, connectives and phrases from English are used in combination with other languages; this can very well be seen in the social media text and in spoken conversations across geographies, especially in India. 

Sentiment Analysis in social media has drawn attention in recent years. However, sentiment analysis on Tamil-English (Tanglish) and Malayalam-English code-mixed data are not readily available for research. The authors of \cite{chakravarthi-etal-2020-corpus} and \cite{chakravarthi-etal-2020-sentiment} have collected 184,573 sentences for Tamil and  116,711 sentences for Malayalam from YouTube comments which are based on the trailers of the movies released in 2019 for building Tamil-English and Malayalam-English datasets where non-code-mixed sentences were removed from the collection. Further, emoticons were removed and sentence length filters were applied to render the mentioned datasets. In the end two data sets of size 15,744 and 6,738 sentences were reported for Tanglish and Malayalam-English texts.

To get this dataset ready for sentiment analysis \cite{chakravarthi-etal-2020-corpus} and \cite{chakravarthi-etal-2020-sentiment} refers to manual annotation activity carried out with three annotators annotating each sentence in the data set. The Krippendorff’s alpha (α) is used to measure inter-annotator agreement which is 0.6585 and 0.890 for Tamil and Malayalam code-mixed data sets respectively. 

\begin{table}
    \centering
    \caption{Dataset size and splits}
    \begin{tabular}{|c|c|c|c|}
         \hline
         Dataset & Training & Validation & Testing \\
         \hline
         Tamil-English & 1,335 & 1,260 & 3,149\\
         Malayalam-English & 4,716 & 674 & 1,348 \\
         \hline
    \end{tabular}
    \label{tab:Dataset_size}
\end{table}

The dataset was provided for this task in three parts training, validation and testing. The number of sentences used for the dataset splits are provided in table \ref{tab:Dataset_size}. Both these datasets are released in DravidianCodeMix FIRE 2020 competition organized by dravidiancodemixed. These comments were grouped into five categories positive, negative, neutral, mixed emotions, or not in the intended languages. 

\section{Methodology}

Language models have been integral to the recent advances made in the field of NLP due to its ability to predict the next token in a sequence. Traditionally this achieved by computing the joint distribution of the tokens in a sequence as a function of conditional probability distribution of each token given other tokens in the sequence.

However, XLNet\cite{yang2019xlnet} takes a different approach; when these models are trained on large datasets, it achieves state-of-art performances on downstream NLP tasks. This uses permutation language modelling, which trains an autoregressive model on all possible permutation of words in a sentence -- see equation \ref{eq:xlnet}. During prediction of a word in a sequence, it takes into account bidirectional context and predicts the masked tokens on the basis of the words/tokens to the right as well as the left of the masked token in the sequence. XLNet is based on the transformer architecture\cite{vaswani2017attention}, which uses the concept of attention\cite{bahdanau2014neural} to learn the long range token dependencies. Another important aspect of XLnet is two-stream attention; this refers to attention streams working in parallel, one which encodes the content of the tokens and the other which incorporates the positional information. This property would be useful and so is exploited to perform the sentiment analysis on code-mixed data.

The following equation formally describes the language modelling objective using XLNet. In eqn.\ref{eq:xlnet}, for a give text sequence x, and set of all permutations of the sequence ${Z_T}$ and z ${\epsilon}$  ${Z_T}$.  

\begin{equation}
\underset{\theta} max \: E_{z\sim {Z_T}}[\sum_{t=1}^{T} {\log p}_{\theta}\:x_{z_t}|x_{z<t})\:] 
\label{eq:xlnet}
\end{equation}


For the purpose of experiments, we fine-tuned the XLNet model using the given datasets -- refer to table \ref{tab:Dataset_size} and sentiment analysis was conducted on this labelled dataset. The training and testing were carried out as per the numbers mentioned in Table \ref{tab:Dataset_size}. Two experiments were conducted on the mentioned datasets and for those experiments the XLNet embeddings were fine-tuned for 4 epochs each with a maximum learning rate of 0.005 to perform sentiment analysis of the given datasets. Results of the experiments carried out are illustrated in the next section. 

\begin{table}
\caption{Precision, Recall and F-score measures on the Test set}
\smallskip\noindent
\resizebox{\linewidth}{!}{%
\begin{tabular}{ |c|c|c|c|c|c|c|} 
\hline
Data & Classes & Precision & Recall & F1 Score & Weighted Average-F1 & Accuracy \\
\hline
\multirow{5}{*}{Malayalam-English} & Mixed feelings & 0.03 & 0.22 & 0.05 & \multirow{5}{*}{0.52} & \multirow{5}{*}{ 0.49 }\\
& Negative & 0.12 & 0.40 & 0.19&&\\ 
& Positive & 0.72 & 0.50 & 0.59&&\\
& not-malayalam & 0.36 & 0.58 & 0.44&&\\
& unknown state & 0.42 & 0.46 & 0.44&&\\
\hline
\multirow{5}{6em}{Tamil-English} & Mixed feelings & 0.23 & 0.13 & 0.17 & \multirow{5}{3em}{0.32} & \multirow{5}{3em}{ 0.35 }\\ 
& Negative & 0.50 & 0.16 & 0.24&&\\ 
& Positive & 0.39 & 0.73 & 0.51&&\\
& not-Tamil & 0.10 & 0.40 & 0.16&&\\
& unknown state & 0.03 & 0.31 & 0.05&&\\
\hline
\end{tabular}}
\label{tab:Results}
\end{table}

\section{Results and Discussion}
The experiment results for the experiments outlined in the previous section are provided in table \ref{tab:Results}. We were able to achieve 0.49 \& 0.35 accuracies and 0.52 \& 0.32 F-scores on both the datasets respectively. The results are biased towards {\em Positive} class because of the class-imbalance seen in the training set. 
Further, it can be seen that the model performs better on the Malayalam-English dataset despite the Tanglish dataset having more samples; this can be attributed to more noise in the Tamil-English data and hence relatively poor performance. Our results do not perform better than the baseline-results described in \cite{chakravarthi-etal-2020-corpus} and \cite{chakravarthi-etal-2020-sentiment}. We hypothesize that these results can further be improved by training the model for more epochs with a pre-processing step performed in combination with oversampling and undersampling of the minority and majority classes respectively.

\bibliography{sample-ceur}

\end{document}